# A Genetic algorithm to solve the container storage space allocation problem

I. Ayachi [*,**], R. Kammarti [*,**], M. Ksouri [**] and P. Borne [*]

[*] LAGIS, Ecole Centrale de Lille, Scientific city - BP 48 - 59651 Villeneuve d'Ascq Cedex - France

[**] LACS, Ecole Nationale des Ingénieurs de Tunis – BP 37, Le Belvédère, 1002 Tunis - Tunisia

ayachiimen@gmail.com , kammarti.ryan@planet.tn , Mekki.Ksouri@insat.rnu.tn, pierre.borne@ec-lille.fr

*Abstract* - This paper presented a genetic algorithm (GA) to solve the container storage problem in the port. This problem is studied with different container types such as regular, open side, open top, tank, empty and refrigerated containers. The objective of this problem is to determine an optimal containers arrangement, which respects customers' delivery deadlines, reduces the re-handle operations of containers and minimizes the stop time of the container ship.
In this paper, an adaptation of the genetic algorithm to the container storage problem is detailed and some experimental results are presented and discussed. The proposed approach was compared to a Last In First Out (LIFO) algorithm applied to the same problem and has recorded good results.

## I. INTRODUCTION

The container storage problem can be defined as a bin packing problem in three dimensions where containers are items and storage spaces in the port or in the ship are bins used. It falls into the category of NP hard problems. At each port of destination, some containers are unloaded from the container ship and loaded in the port to be delivered to their customers. In fact, a container is either in the vessel waiting her unloading in the destination port or in at the harbor port waiting to be loaded in the vessel or to be delivered to his customer.

Most of previous works studied the container storage problem in the ship. There objective was to determine an optimal container arrangement that satisfy there criteria such the ship stability, the reduction of the re-handle operations of containers, the minimization of the total loading time …

Many approaches have been developed to solve this problem, rule based, mathematical model, simulation based, multi-agent model [2] and heuristic methods. Nevertheless, the temporary storage of containers at the container terminal is one of the most important services. It affects directly on the efficiency of the port, the handling equipment and consequently the transportation costs.This problem is known as the storage space allocation problem (SSAP). It consists on affecting each container to the most suitable storage area in accordance with the problem objective.

The main objective of this paper is to solve the container storage allocation using genetic algorithm. Our aim is to determine a valid containers arrangement in the port, in order to respect customers' delivery deadlines, reduce the loading/unloading times of these containers as well as their re-handle operations.

The contribution of this work is that it solves the problem with different containers types (dry, open side, open top, tank, empty and refrigerated). Indeed, many storage constraints appear linked to this diversity type such as refrigerated containers must be allocated to the blocks equipped by the power point, open top containers can not have another container at the top, tank container must be placed on each other, ... So, the problem becomes more complicated.

In this paper, a genetic algorithm is proposed to solve the container storage space allocation. This approach is chosen since his facility and quick achieve to the feasible solutions even for models having numerous equality constraints. [1],[3]. A comparative study between the proposed approach and the LIFO algorithm was performed.

## II. LITERATURE REVIEW

Many approaches have been developed to solve the storage space allocation problem: simplified analytical calculations or detailed simulation studies. In the literature [4], different metaheuristics (tabu search, simulated annealing and genetic algorithms) were combined to solve the port yard storage optimization problem (PYSOP). The problem is akin to a two dimensional Bin packing problem aims to minimize the space allocated to the cargo within a time interval.

In the paper [5], authors present a simple analytical model for predicting unloading containers times and determining equipment utilization. The prediction model was applied in the Suva's port and has recorded encouraging results.

In the literature [9], a new metaheuristic called harmony search was developed to solve the SSAP. The proposed approach was compared to a genetic algorithm [8] previously applied to the same problem and recorded a good results.

Zhang and his colleagues [6] solved the SSAP using a rolling-horizon approach. Their aim was to minimize the total distance to transport the containers. For each planning horizon, they decomposed the problem into two levels and formulated each level as a mathematical programming model. In the work [1], we have an extended of the SSAP proposed in the literature [6] when the type and the size of containers are different. Bazzazi and his colleagues used the genetic algorithm to solve this problem and they supposed that allowable blocks that a container type can be allocated to them are known in advance.

Due to encouraging results reported in this paper, a genetic algorithm is developed in this work. His aim is to find an optimal solution that stored all containers when the type is different, respects all constraints equations and customers' delivery deadlines and reduces the re-handling operations.

### III. PROBLEM FORMULATION

In this section, we detail our evolutionary approach by presenting the adopted mathematical formulation and the evolutionary algorithm based on the following assumptions.

#### A. Assumptions

In our work we suppose that:

- Initially containers are unloaded from the vessel and transmitted to storage area waiting to be allocated to the allowable place in the storage block.
- To unload a container, all containers above must be re-handled.
- Each container is waiting to be delivered to their customer.
- The containers are of different types (dry, open top, open side, tank, empty and refrigerated) and have the same size.

The storage area in the port is composed of a several blocks which can be equipped by a power point to stored reefer containers or regular blocks for the other type of containers. Figure 1 shows an example of storage area.

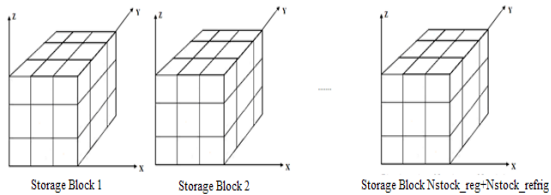

Fig.1 Storage area

#### B. Input parameters

Let's consider the following variables:

- *i:* Container index,
- $n_1$: Maximum containers number on the axis X
- $n_2$: Maximum containers number on the axis Y
- $n_3$: Maximum containers number on the axis Z
- $Nc_{floor}$: Maximum containers number per floor, $Nc_{floor} = n_1 * n_2$
- $Nc_{floor}(k)$ : the containers number in the floor k
- $N_{stock\_reg}$ : the storage blocks number for container don't required power point.
- $N_{stock\_refrig}$ : the storage blocks number for refrigerated containers.
- *j:* storage block index, j=1,…, $N_{stock\_reg}$ + $N_{stock\_refrig}$
- $d_i$: delivery date of each container i

- *Nc(T):* the containers number for each type, with T denotes the container type, T=1,2,…6

$$T = \begin{cases} 1 & \text{if it's a dry container} \\ 2 & \text{if it's an empty container} \\ 3 & \text{if it's an open top container} \\ 4 & \text{if it's an open side container} \\ 5 & \text{if it's a tank container} \\ 6 & \text{if it's a reefer container} \end{cases}$$

- $N_T$: number of container type
- $Nc_{max}$: Maximum containers number, with
  $Nc_{max} = (N_{stock\_reg} + N_{stock\_refrig}) n_1.n_2.n_3$

#### C. Decision Variable

$Pos(x, y, z, j)$ is the decision variable. It's a structure with two variables such as V and Typ.

Where: $x \in [1,..,n_1]$, $y \in [1,..,n_2]$, $z \in [1,..,n_3]$, $j \in [N_{stock\_gene} + N_{stock\_refrig}]$

$$Pos(x,y,z,j).V = \begin{cases} 1 & \text{if we have a container in this position} \\ 0 & \text{otherwise} \end{cases} \quad (1)$$

$$Pos(x,y,z,j).typ = \begin{cases} T & \text{if we have a container in this position} \\ 0 & \text{otherwise} \end{cases} \quad (2)$$

#### D. Mathematical formulation

Let us consider that the storage area at the port consisting of a defined blocks number. Our fitness function aims to reduce the number of container rehandlings and then minimize the ship stoppage time. To do that, the following function is used.

*Fitness function*

$$\text{Minimise} \sum_{t=1}^{N_T} \sum_{i=1}^{Nc(T)} \sum_{j=1}^{N_{stock\_reg}+N_{stock\_refrig}} P_{i,t} m_{i,t,j} Pos(x,y,z,j).V$$

$$\forall x = 1..n_1, \forall y = 1..n_2, \forall z = 1..n_3$$

Where:

- $P_i$ : Priority value depending on the delivery date $d_i$ of container i to customer, with $P_i = 1/d_i$
- $m_{i,j}$: the minimum number of container rehandles to unload the container i which is in the storage block j

#### E. Constraints:

The constraint equations (3) and (4) ensure that a floor lower level contains more containers than directly above.

$$Nc_{floor}(k) \geq Nc_{floor}(k+1), k = 1,…,N_{floor} \quad (3)$$

$$\text{If } Pos(x,y,z,j) = 0 \text{ Then } Pos(x,y,z-1,j) = 0 \quad (4)$$

Constraint (5) indicates that an open top container can not have another container above.

$$\text{If } \begin{cases} Pos(x,y,z,j).V = 1 \\ \text{and} \\ Pos(x,y,z,j).typ = 3 \end{cases} \text{Then } Pos(x,y,z+1,j).V = 0 \quad (5)$$

Constraint (6) ensures that an open side container

$$\text{If } \begin{cases} Pos(x,y,z,j).V = 1 \\ \text{and} \\ Pos(x,y,z,j).typ = 4 \end{cases} \text{Then } \begin{cases} Pos(x,y,z+n,j).V = 0 \\ \forall n \in [1,..,n_3 - z[ \\ Pos(x+m,y,z,j).V = 0 \\ \forall m \in [1,..,n_1 - x[ \end{cases} \quad (6)$$

Constraint (7) ensures that an empty container must not be settled under a full container.

$$\text{If} \begin{cases} Pos(x,y,z,j).V = 1 \\ \text{and} \\ Pos(x,y,z,j).typ = 2 \end{cases} \text{Then} \begin{cases} Pos(x,y,z+1,j).V = 0 \\ \text{Or} \\ Pos(x,y,z+1,j).typ = 2 \end{cases} \quad (7)$$

Constraint (8) suggests that tank containers must be placed on each other.

$$\text{If} \begin{cases} Pos(x,y,z,j).V = 1 \\ \text{and} \\ Pos(x,y,z,j).typ = 5 \end{cases} \text{Then} \begin{cases} Pos(x,y,z+1,j).typ = 5 \\ \text{Or} \\ Pos(x,y,z+1,j).V = 0 \end{cases} \quad (8)$$

Constraint (9) indicates that a reefer container must be allocated to the blocks equipped by the power point.

$$\text{If} \begin{cases} Pos(x,y,z,j).V = 1 \\ \text{and} \\ Pos(x,y,z,j).typ = 6 \end{cases} \text{Then } j \in [1,..,N_{stock\_refrig}] \quad (9)$$

## IV. EVOLUTION PROCEDURE

In this section, genetic algorithm implementation is detailed. The principle of the selection procedure is the same used by Kammarti in [7] and [8].

Initially, a population of size N was created randomly. Then, two parents were selected using roulette-wheel method and N new solutions generated using the two-point crossover operator and mutation. The new population added to the initial to form an intermediate population noted $P_{inter}$ and having 2N as size. $P_{inter}$ is sorted according to their fitness in increasing order. The first N individuals of $P_{inter}$ will form the population (i +1), where i is the iteration number. This procedure will be repeated until the termination criterion (number of improvisations) is satisfied.

### A. Solution representation chromosome

According to the decision variable Pos(x, y, z, j), we use a four dimension structure representation, witch reproduce the containers storage area. These dimensions indicate respectively the container coordinates in the plan (X, Y, Z) and the number of the allocated block.

### B. Initial population generation procedure

The initial population is randomly generated, where every stored solution must respect all problem constraints (equations (1) to (9)).

### C. Crossover operator

The crossover operator adopted consists on choosing two parents I1 and I2 from the initial population using roulette-wheel selection and a randomly crossover plan defined ($p_{crois-x}$, $p_{crois-y}$, $p_{crois-z}$). The crossover operation will be produced with a probability fixed to 70%

### D. Mutation operator

The mutation operator consists of permuting two randomly selected containers. The mutation probability is set to 20%.

## V. EXPERIMENTAL RESULTS

In this section, the performance of the proposed genetic algorithm is assessed for different simulations. For the proposed approach, the algorithm stops when the solution doesn't improve after $N_{iter}$ iterations. In addition, it is supposed that:

- $n_1$, $n_2$ and $n_3$ will be defined by the user.
- The containers type $N_T$, the number of each container type Nc(T) and the storage blocks number ($N_{stock\_refrig}$, $N_{stock\_reg}$) are defined by the user.
- The delivery date of each container is randomly defined.

To evaluate the results of the proposed genetic algorithm, the influence of the container type number, the stopping criteria and the population size are studied.

### A. The number of containers type influence

To study the influence of the number of container types, the algorithm is executed for different values of $N_T$. For each simulation, the best fitness values of the first ($F_i$) and the last ($F_f$) iterations are given and the execution time ($T_{Execution}$) is calculated.

To do so, the population size was set to 50, the stopping criteria ($N_{iter}$) to 20, $n_1 = n_2 = n_3 = 3$, $N_{stock\_reg}$=4 and $N_{stock\_refrig}$= 4. The results are presented in table I.

TABLE I.
CONTAINER TYPE INFLUENCE

| $N_T$ | $N_c(T)$ | $F_i$ | $F_f$ | $T_{Execution}$ (s) |
|---|---|---|---|---|
| 2 | $N_c(1)$=10,$N_c(2)$=10 | 1,92 | 0 | 8 |
| 3 | $N_c(1)$=10, $N_c(2)$=10 $N_c(3)$=8 | 2,82 | 0 | 11 |
| 4 | $N_c(1)$=10, $N_c(2)$=10 $N_c(3)$=8,$N_c(4)$=8 | 21,38 | 0 | 14 |
| 5 | $N_c(1)$=10, $N_c(2)$=10 $N_c(3)$=8, $N_c(4)$=8 $N_c(5)$=15 | 39,45 | 0 | 18 |
| 6 | $N_c(1)$=10, $N_c(2)$=10 $N_c(3)$=8, $N_c(4)$=8 $N_c(5)$=15, $N_c(6)$=10 | 54,81 | 0 | 27 |

As it can be seen, higher is the containers number, harder is the problem. In addition, it becomes more complex and its execution time increases. This can be explained by the appearing of diverse constraints related to the arrangement conditions of each container type.

### B. The stopping criteria value influence

In order to study the influence of the stopping criteria parameter, we varied $N_{iter}$ and we fixed the following values:

- The container type $N_T = 4$
- The population size to 30, $n_1 = n_2 = n_3 = 3$, $N_{stock\_reg}$=3, $N_{stock\_refrig}$= 3

TABLE II.
STOPPING CRITERIA INFLUENCE

| $N_{iter}$ | $F_i$ | $F_f$ | $T_{Execution}$(s) |
|---|---|---|---|
| 25 | 33,24 | 11,13 | 8 |
| 50 | 49,47 | 10,89 | 17 |
| 100 | 34,56 | 9,39 | 21 |
| 150 | 41,98 | 8,08 | 31 |

According to the results illustrated in Table II, we note that higher is the value of the stopping criteria, better is the quality of the fitness function.

However, the execution time increases with the stopping criteria value.

*C. The population size influence*

Through this example, the size of the problem is fixed to 4 types of containers (dry, open side, tank, reefer) with Nc(1)= 20, Nc(4)=15, Nc(5)= 10, Nc(6)=30 and $N_{iter}$ to 50 iterations. The population size N, is varied to study his influence on the algorithm behavior. The results are presented in the following table.

TABLE III.
POPULATION SIZE INFLUENCE

| N | Fi | Ff | $T_{Execution}$ (s) |
|---|---|---|---|
| 20 | 27,70 | 13,68 | 9 |
| 40 | 24,95 | 11,16 | 14 |
| 50 | 21,34 | 10,36 | 17 |
| 70 | 20,93 | 9,72 | 24 |
| 100 | 19,76 | 8,54 | 31 |

The results shown in the table III indicate that higher is the population size, better is the value of the fitness function. This may be explained by the fact that the increase of the population size should allow a better exploration and exploitation of the feasible space

## VI. COMPARATIVE STUDY

To evaluate the results generated by the genetic algorithm proposed in this paper, a comparative study with a LIFO algorithm was performed.

TABLE IV.
DIFFERENT STUDIED CASES DESCRIPTION

| Instance N° | $N_T$ | $N_c(T)$ |
|---|---|---|
| 1 | 2 | $N_c(1)$=50 $N_c(3)$=15 |
| 2 | 3 | $N_c(1)$=25 $N_c(2)$=25, $N_c(3)$=10 |
| 3 | 4 | $N_c(3)$=8, $N_c(4)$=5, $N_c(5)$=7, $N_c(6)$=15 |
| 4 | 5 | $N_c(2)$=14, $N_c(3)$=8, $N_c(4)$=5, $N_c(5)$=7 $N_c(6)$=15 |
| 5 | 6 | $N_c(1)$=25 $N_c(2)$=14, $N_c(3)$=9, $N_c(4)$=8 $N_c(5)$=7, $N_c(6)$=12 |

The LIFO algorithm consists on storing in first time the last placed container in a stack. This principle is applied in most port container terminals, where a manual planning based on experience and rules to assign each container to a certain storage block. The performance of the two approaches is verified according to the 5 case studies described in table IV, by varying the containers numbers and types.

Each case study is solved using the genetic algorithm 15 times and the mean of fitness values and execution times are calculated. For these experimentations the population size is set to 30, $N_{iter}$ to 20, $n_1$, $n_2$ and $n_3$= 3, $N_{stock\_reg}$ to 3 and $N_{stock\_refrig}$ to 2.

As it can be seen in table V, the solutions generated by the GA are largely better from the ones generated by the LIFO technique, especially when the problem size grows. However, the LIFO approach is faster.

TABLE V.
COMPARISON BETWEEN LIFO AND GA'S FITNESS VALUE AND EXECUTION TIME

| Instance N° | LIFO Algorithm | | Genetic algorithm | |
|---|---|---|---|---|
| | Fitness value | $T_{Execution}$ (s) | Fitness value | $T_{Execution}$ (s) |
| 1 | 3,65 | 0,5 | 0 | 20 |
| 2 | 5,59 | 2 | 0 | 22 |
| 3 | 4,72 | 4 | 0 | 37 |
| 4 | 10,14 | 4,5 | 1,29 | 65 |
| 5 | 19,37 | 6 | 3,16 | 80 |

## VII. CONCLUSION

In this paper, a genetic algorithm is proposed to solve the container storage problem in the port. The objective aims to determine the best containers arrangement that meet customers' delivery dates and reduce the number of container rehandlings.

The contribution of this work is that it solves the problem with different containers types (dry, open side, open top, tank, empty and refrigerated).

The type of container must be considered on the allocation of containers to the storage blocks. Since some storage constraints must be respected for each type. The proposed approach has provided encouraging results when compared to the LIFO algorithm used by manual planner in the harbor.